\documentclass[runningheads]{llncs}
\usepackage{graphicx}
\usepackage{color}

\usepackage{xcolor}
\usepackage{tabularx}
\usepackage{ragged2e}
\usepackage{makecell}
\usepackage{amsmath}
\usepackage{amssymb}

\usepackage[pagebackref=true,breaklinks=true,letterpaper=true,colorlinks,bookmarks=false]{hyperref}

\begin{document}
\title{Insights From A Large-Scale Database of Material Depictions In Paintings}
%
%

\author{Hubert Lin$^{*}$\inst{1} \and Mitchell Van Zuijlen\inst{2} \and Maarten W.A.
Wijntjes\inst{2} \and Sylvia C. Pont\inst{2} \and Kavita Bala\inst{1} }
\authorrunning{H. Lin et al.}
%
\institute{Cornell University, Computer Science Department \\
\and Delft University
of Technology, Perceptual Intelligence Lab \\
$^{*}$\email{hubert@cs.cornell.edu} \\
}

\maketitle              
\begin{abstract} 

  Deep learning has paved the way for strong recognition systems which are often
  both trained on and applied to natural images. In this paper, we examine the
  give-and-take relationship between such visual recognition systems and the
  rich information available in the fine arts. First, we find that visual
  recognition systems designed for natural images can work surprisingly well on
  paintings. In particular, we find that interactive segmentation tools can be
  used to cleanly annotate polygonal segments within paintings, a task which is
  time consuming to undertake by hand. We also find that FasterRCNN, a model
  which has been designed for object recognition in natural scenes, can be
  quickly repurposed for detection of materials in paintings. Second, we show
  that learning from paintings can be beneficial for neural networks that are
  intended to be used on natural images.  We  find that training on paintings
  instead of natural images can improve the quality of learned features and we
  further find that a large number of paintings can be a valuable source of test
  data for evaluating domain adaptation algorithms.  Our experiments are based
  on a novel large-scale annotated database of material depictions in paintings
  which we detail in a separate manuscript.

\end{abstract}

\section{Introduction}

Deep learning has enabled the development of high performing recognition systems
across a variety of image-based tasks
\cite{garcia2018survey,hafiz2020survey,ye2020deep}.  These systems are often
trained on natural photographs with applications in real world recognition like
self-driving. Furthermore, applying recognition systems to large collections of
images can also reveal cultural trends or give us insight into the visual
patterns in the world (e.g.
\cite{matzen2017streetstyle,mall2019geostyle,lin2020visual}).  Human-created
images, such as paintings, are particularly interesting to analyze from  this
perspective.  Artistic depictions can reveal insights into culturally relevant
ideas throughout time, as well as insights into human visual perception through
the realism depicted by skilled artists. 

Whereas most computer vision systems focusing on digital art history are
concerned with \emph{object} recognition (e.g., \cite{Crowley14}), it is the
depiction of \emph{space} and \emph{materials} that visually characterized the
course of art history.  The depiction of space has had considerable attention in
scientific literature
\cite{panofsky2020perspective,white1957birth,kemp1990science,pirenne1970optics}
while recently the depiction of materials has gained scientific interest
\cite{beurs,wiersma2020,pottasch2020,wijntjes2020}. Therefore, it is interesting
to investigate the interplay between deep learning systems designed for natural
image analysis and the rich visual information found in paintings, especially
with respect to artistic depictions of materials. 

The remainder of this paper is organized into three parts. In Section 2, we
briefly describe the dataset that subsequent experiments are based on.  In Section 3,
we explore how deep learning systems that have  primarily been developed for use
on natural photographs can be used to analyze paintings.  Specifically, we
explore (a) segmentation and (b) detection of materials in paintings.
Recognition of materials in paintings can be useful for digital art history as
well as general public interest. In Section 4, we explore how paintings can be a
useful source of data from which better recognition systems can be built.
Specifically, we investigate (c) the generalizability and interpretability of
classifiers trained on paintings, and we investigate (d) the role that a
large-scale painting dataset can play in evaluating visual recognition models.

\section{Dataset}

\begin{figure}[h!]
    \begin{center}
    \includegraphics[width=0.7\linewidth]{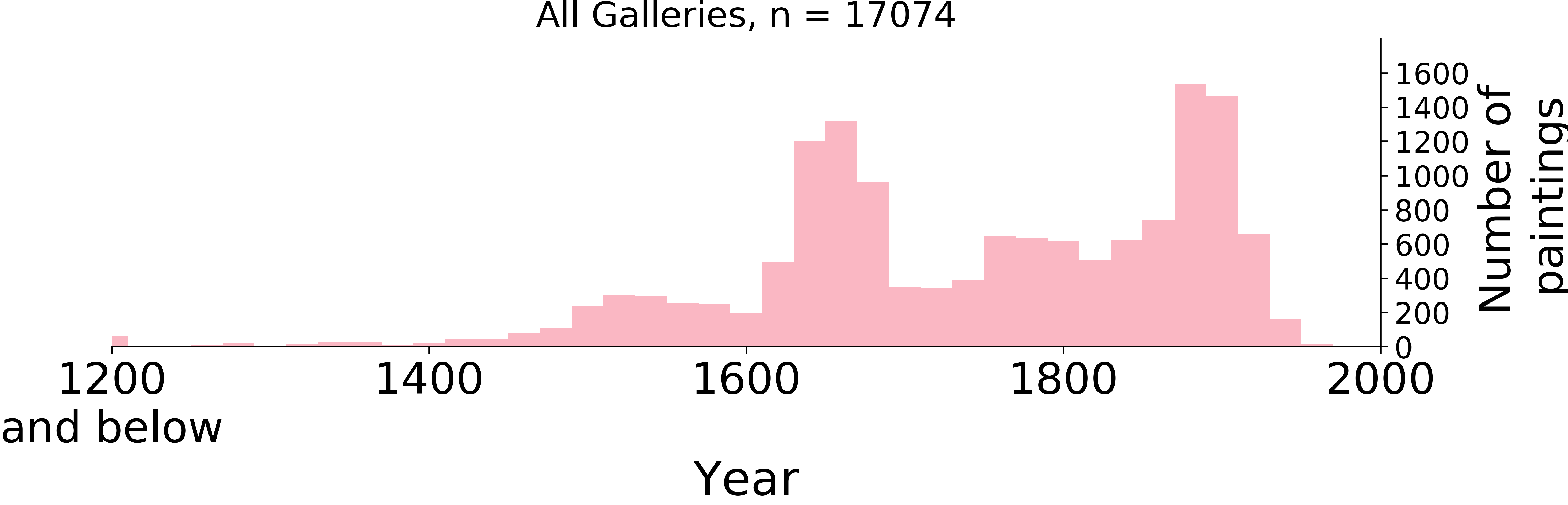}
      \vspace{-4mm}

      \caption{ \textbf{Year Distribution of Paintings in Dataset.} Each bin
      equals 20 years. There are peaks in the paintings in the 1700s and 1900s.
      The former corresponds to the European golden ages; it is less clear what
      explains the latter peak.}

      \label{fig:year}
    \end{center}
\end{figure}

\begin{figure}[h!]
    \begin{center}
      \includegraphics[width=0.7\linewidth]{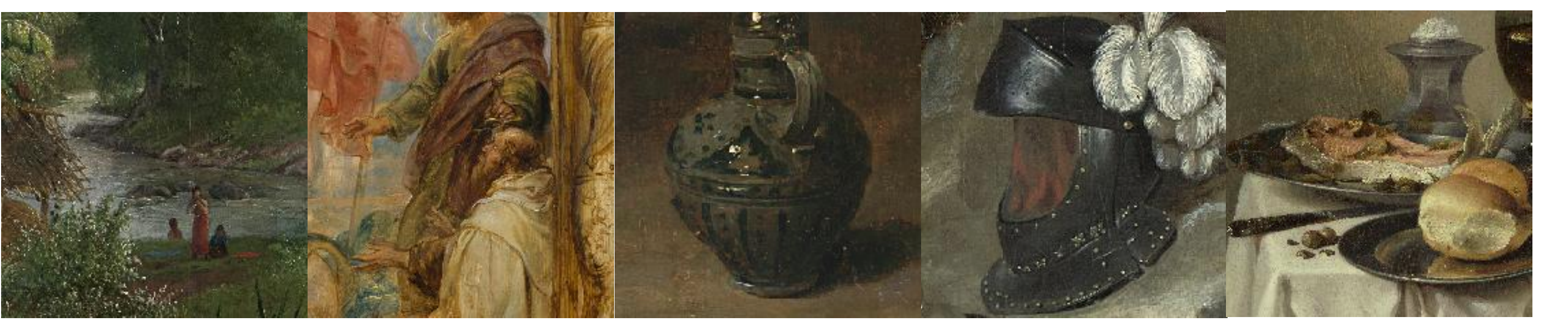}
      \vspace{-4mm}
      \caption{ \textbf{Examples of Annotated Bounding Boxes.} Left to
      Right: Liquid, Fabric, Ceramic, Metal, and Food.}
      \label{fig:bb_ex}
    \end{center}
\end{figure}

All experiments in this paper utilize data from the Materials in Paintings (MIP) dataset,  a large-scale annotated dataset
of material depictions in paintings. Extensive details and analysis of this
database will be available in a separate manuscript. For context and
completeness, we summarize a few relevant details here. The dataset consists of
19K high resolution paintings downloaded from the online collections of
international art galleries, which span over 500 years of art history. The
galleries with corresponding number of paintings are: The Rijksmuseum (4,672),
The Metropolitan Museum of Art (3,222), Nationalmuseum (3,077), Cleveland Museum
of Art (2,217), National Gallery of Art (2,132), Museo Nacional del Prado
(2,032), The Art Institute of Chicago (936), Mauritshuis (638), and J. Paul
Getty Museum (399). The distribution of paintings by year is shown in Fig.
\ref{fig:year}. The dataset includes crowdsourced extreme click
\cite{Papadopoulos2017} bounding box annotations over 15 material categories,
which are further delineated into 50 finegrained categories. Fig.
\ref{fig:bb_ex} shows a few examples of the annotated bounding boxes available
in the dataset.

\section{Using Computer Vision to Analyze Paintings}

Research in computer recognition systems have focused primarily on natural images.
For example, semantic segmentation benchmarks (of objects, `stuff', or
materials) \cite{openimages,lin2014,pascalvoc,cocostuff,opensurfaces,minc} emphasize parsing
in-the-wild photos, with applications in robotics, self-driving, and so forth.
However, the analyses of paintings can also benefit from the use of visual
recognition systems. Paintings can encode both cultural and perceptual biases,
and being able to analyze paintings at scale can be useful for a variety of
scientific disciplines including digital art history and human visual perception.

In this section, we explore the effectiveness of interactive segmentation
methods (which can be used to select regions of interest in photographs for
the purpose of image editing or data annotation) when applied to paintings. We
also explore how well an \emph{object} bounding box detector can be finetuned to detect
\emph{materials} depicted in unlabelled paintings, which could be used for
content-based retrieval.

\subsection{Extracting Polygon Segments with Interactive Segmentation}

Polygon segmentation masks are useful for reasoning about boundary relationships
between different semantic regions of an image, as well as the shape of the
regions themselves.  However, annotating segmentations is expensive and many
modern datasets rely on expensive manual annotation methods
\cite{opensurfaces,lin2014,ade20k,cocostuff,cityscapes}. Recent work has focused
on more cost effective annotation methods (e.g.
\cite{lin2019block,Maninis2017,benenson2019large,ling2019fast}). The use of
interactive segmentation methods that transform sparse user inputs into
polygon masks can ease annotation difficulty. For paintings, it is unclear whether
these methods (especially deep learning methods trained on natural images) would
perform well.  Semantic boundaries in paintings likely have a different, and
more varying structure than in photos. Paintings can have ambiguous or fuzzy
boundaries between objects or materials \cite{hommes} which can potentially be
problematic for color-based methods. This can be due to variations in artistic
style which can emphasize different aspects of depictions -- for example, Van
Gogh uses lines and edges to create texture, but such edges could potentially
appear as boundaries to a segmentation model. In this experiment, we study the
difficulty of segmenting paintings and whether innovations are necessary for
existing methods to perform well.

\subsubsection{Experimental Setup.} We experiment with GrabCut \cite{grabcut}
(an image-based approach) and DEXTR \cite{Maninis2017} (a modern deep learning
approach). We evaluated the performance of these methods against ~4.5k
high-quality human annotated segmentations from \cite{van2020painterly}. The
inputs to these methods are generated from the extreme points of the regions we
are interested in. We use a variant of the GrabCut initialization proposed in
\cite{Papadopoulos2017}, as well as a rectangular initialization for reference. For
DEXTR, we consider models pretrained on popular object datasets
\cite{lin2014,pascalvoc} as a starting point.

\subsubsection{Results.} 

\begin{figure}[ht!]
    \centering
    \includegraphics[width=0.7\linewidth]{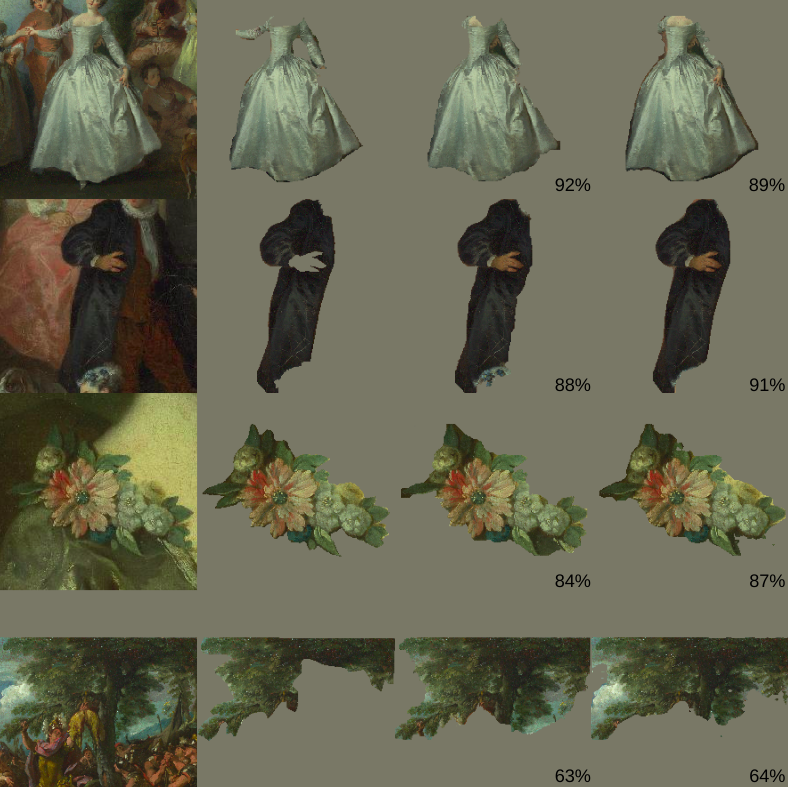}

    \vspace{-4mm}
    \caption{ \textbf{Extreme Click Segmentations.} Left to right: Original
    Image, Ground Truth Segment, Grabcut Extr Segment, DEXTR COCO Segment. Both
    Grabcut and DEXTR use extreme points as input. For evaluation, the extreme
    points are generated synthetically from the ground truth segments. In
    practice, extreme clicks can be crowdsourced. Bottom-right corner shows the
    IOU for each segmentation.}

    \label{fig:segments}
\end{figure}

We found that both GrabCut and DEXTR  perform quite well on paintings.
Surprisingly, DEXTR transfers quite well to materials in paintings despite being
trained only with natural photographs of objects. The performance of DEXTR can
be further improved by finetuning on COCO with a smaller learning rate (10\% of
original learning rate for 1 epoch). Finetuning DEXTR on Grabcut segments or
iteratively finetuning with output of DEXTR does not seem to yield further
improvements.  The performance is summarized in Table \ref{table:segments}, and
samples are visualized in Fig. \ref{fig:segments}.  

\begin{table}[ht!] 
  \centering
  Segmentation mean IOU (\%)\\ 
  \begin{tabularx}{0.61\linewidth}{|c|c|c|c|c|}
    \hline
    \makecell[c]{Grabcut\\Rectangle} & \makecell[c]{Grabcut\\Extr} &
  \makecell[c]{DEXTR\\Pascal-SBD} & \makecell[c]{DEXTR\\COCO} &
  \makecell[c]{DEXTR\\Finetuned}\\ 
    \hline 
    44.1 & 72.4 & 74.3 & 76.4 & 78.4 \\
  \hline 
  \end{tabularx} 
  \vspace{2mm} \\ 
  DEXTR Finetuned IOU By Class (\%)\\
  \begin{tabularx}{0.46\linewidth}{|c|c|c|c|c|} 
    \hline Animal &  Ceramic &
    Fabric & Flora & Food   \\
    \hline 
    76.9 & 86.8 & 79.1 & 77.0 & 87.5 \\
    \hline
    \hline
    Gem & Glass & Ground & Liquid & Metal \\
    \hline 
    74.4 & 83.2 & 69.6 & 73.0 & 75.5 \\
    \hline
    \hline
    Paper & Skin & Sky & Stone & Wood 
    \\ 
    \hline
  86.1 & 78.9 & 78.5 & 81.7 & 67.4 \\
  \hline
  \end{tabularx}

  \caption{ \textbf{Segmentation Performance}.  Grabcut Extr is based on
  \cite{Papadopoulos2017} with small modifications: (a) minimum cost boundary is
  computed with the negative log probability of a pixel belonging to an edge;
  (b) in addition to clamping the morphological skeleton, the extreme points
  centroid and extreme points are clamped; (c) GC is computed directly on the
  RGB image. DEXTR \cite{Maninis2017} is pretrained on Pascal-SBD and COCO.
  Note that Pascal-SBD and COCO are natural image datasets of objects, but DEXTR
  transfers surprisingly well across both visual domain (paintings vs.  photos)
  and annotation categories (materials vs.  objects).} 

\label{table:segments} 
\end{table}

\subsection{Detecting Materials in Unlabeled Paintings}

To allow the public to view and interact with art collections, museums and
galleries provide extensive online functionality to search and navigate through
the collections.  Currently, to our knowledge, no online collections allows
online visitors to query the collection for depicted \emph{materials} within
painting, which can be of interest to the public. Furthermore, depiction of
materials plays a crucial role in characterizing art history. Detecting
materials within novel paintings will be particularly beneficial to digital art
historians who study materials  such as stone
\cite{Steinformen,dietrich1990rocks} or skin
\cite{bol2012painting,lehmann2008fleshing}. Having access to specific materials
can also digital art historians to compare these depictions directly with
respect to painting style or technique.  We experiment with automatic bounding
box detection to ease access to material depictions in unlabelled collections.

\subsubsection{Experimental Setup.} We train a FasterRCNN \cite{fasterrcnn}
bounding box detector to localize and label material boxes with on 90\% of
annotated paintings in the dataset, and evaluate on the remaining 10\%. Default
COCO hyperparameters from \cite{detectron2} are used.  Given the
non-spatially-exhaustive nature of the annotations, many detected bounding boxes
will not be matched against labelled ground truth boxes. However, the dataset is
exhaustively annotated at an image level, and therefore, we report image-level
accuracies.  This can be interpreted as the accuracy of the model in tagging
each image with the types of materials present. The validity of each localized
box can be further quantified through a user study, but we did not perform this
study at this time.

\subsubsection{Results.} 
\begin{figure}[ht!]
    \centering
    \vspace{-10mm}
    \includegraphics[width=0.7\linewidth]{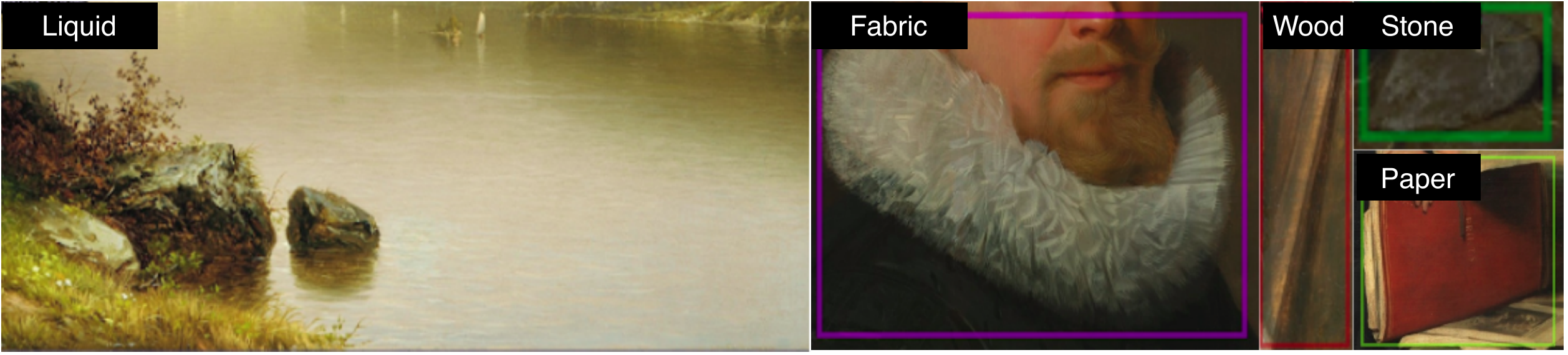}
    \vspace{-4mm}
    \caption{ \textbf{Detected materials in Unlabeled Paintings.}
    Automatically detecting materials can be useful for content retrieval and
    for filtering online galleries by viewer interests.}
    \label{fig:rcnn}
\end{figure}

Table \ref{table:rcnn} shows the performance.  We found
that the FasterRCNN model is able to accurately detect materials in paintings by
finetuning on the annotated bounding boxes directly without any changes to the
network architecture or training hyperparameters. It is certainly promising to
see that an algorithm designed for object localization in natural images can be
readily applied to material localization in paintings.  A qualitative sample of
detected bounding boxes is given in Fig. \ref{fig:rcnn}. To improve the
spatial-specificity of the detected materials, it can be interesting to train an
instance detector like MaskRCNN on segments extracted using methods discussed in
the previous section. It would also be useful to combine material recognition with
conventional object-based detection to extract complementary forms of
information that improve the ability for users to filter data by their specific
needs.

\begin{table}[h!]
    \centering
    \vspace{-5mm}
    Class Accuracy (\%) (Mean = 83.3\%) \\
    \begin{tabularx}{0.46\linewidth}{|c|c|c|c|c|}
    \hline
    Animal & Ceramic & Fabric & Flora & Food   \\
    \hline
    85.6 & 92.7 & 66.0 & 85.0 & 94.9  \\
    \hline\hline
    Gem & Glass & Ground & Liquid & Metal   \\
    \hline
    88.4 & 91.3 & 86.5& 86.4 & 70.7  \\
    \hline\hline
     Paper & Skin & Sky & Stone & Wood   \\
    \hline
    92.4 & 70.2 & 89.4 & 74.8 & 74.9 \\
    \hline
    \end{tabularx}

    \caption{ \textbf{Image-level Detection Accuracy.} Bounding boxes are
    detected with FasterRCNN trained on paintings. Because the dataset is not
    exhaustively annotated spatially, image-level accuracy is reported instead
    of box precision and recall. Overall, images are tagged with the correct
    materials with high accuracy.}

  \label{table:rcnn}
\end{table}

\section{Using Paintings to Build Better Recognition Systems}

In recent work for machine perception systems, art has been used in various
ways.  Models that learn to convert photographs into painting-like or
sketch-like images have been studied extensively for their application as a tool
for digital artists \cite{nstsurvey}.  Recent work has shown that
such neural style transfer algorithms can also produce images that are useful
for training robust neural networks \cite{SIN}. Artworks  have also
been used directly to evaluate the robustness of neural networks under ``domain
shifts'' in which a model trained to recognize objects from photographs are
shown artistic depictions of such objects instead \cite{pacs,domainnet}.

We use the MIP dataset of material depictions in paintings to explore two
directions.  First, we hypothesize that the perceptually focused depictions of
artists can allow neural networks to learn better cues for classification. We
find that learning from paintings can improve the interpretability of the cues
it uses for its predictions. In a second experiment, we investigate the utility
of the MIP dataset as a benchmark for computer vision models
under domain shifts when more data is available than is typical in existing
domain adaptation benchmark datasets. We find that existing domain adaptation
algorithms can fail to behave as expected in this setting.

\subsection{Learning Robust Cues for Finegrained Fabric Classification}

The task of distinguishing between images of different semantic content is a
standard recognition task for computer vision systems. Increasing attention is
being given to ``fine-grained" classification where a model is tasked with
distinguishing images of the same broad category (e.g. distinguishing different
species of birds or different types of flora
\cite{wei2019deep,cub,inaturalist}). Fine-grained classification is particularly
challenging for deep learning systems. Such a task depends on recognizing
specific attributes for each finegrained class; in comparison, classifiers can
perform well on coarse-grained classification by relying on context alone. We
hypothesize that the painted depictions of materials can be beneficial for this
task. Since some artistic depictions focus on salient cues for perception
through perceptual shortcuts \cite{mamassian2008,cavanagh2005a,dicicco2019a}, it
is possible that a network trained on such artwork is able to learn a more
robust feature representation by focusing on these cues.

\subsubsection{Experimental Setup.} We experiment with the task classifying
cotton/wool versus silk/satin.  The latter can be recognized through local cues
such as highlights on the cloth; such cues are carefully placed by artists in
paintings. To understand whether artistic depictions of fabric allow a neural
network to learn better features for classification, we train a model with
either photographs or paintings. High resolution photographs of cotton/wool and
silk/satin fabric and clothing (dresses, shirts) are downloaded and manually
filtered from publicly available photos licensed under the Creative Commons from
Flickr. In total, we downloaded roughly 1K photos. We sample cotton/wool and
silk/satin samples from our dataset to form a corresponding dataset of 1K
paintings.

\paragraph{Generalizability of Classifiers.} Does training with paintings
improve the generalizability of classifiers? To test cross-domain
generalization, we test the classifier on types of images that it has not seen
before. A classifier that has learned more robust features will perform better
on this task than one that has learned to classify images based on more spurious
correlations. We test the trained classifiers on both photographs and paintings.

\begin{figure}[ht!]
    \begin{center}
      \includegraphics[width=0.7\linewidth]{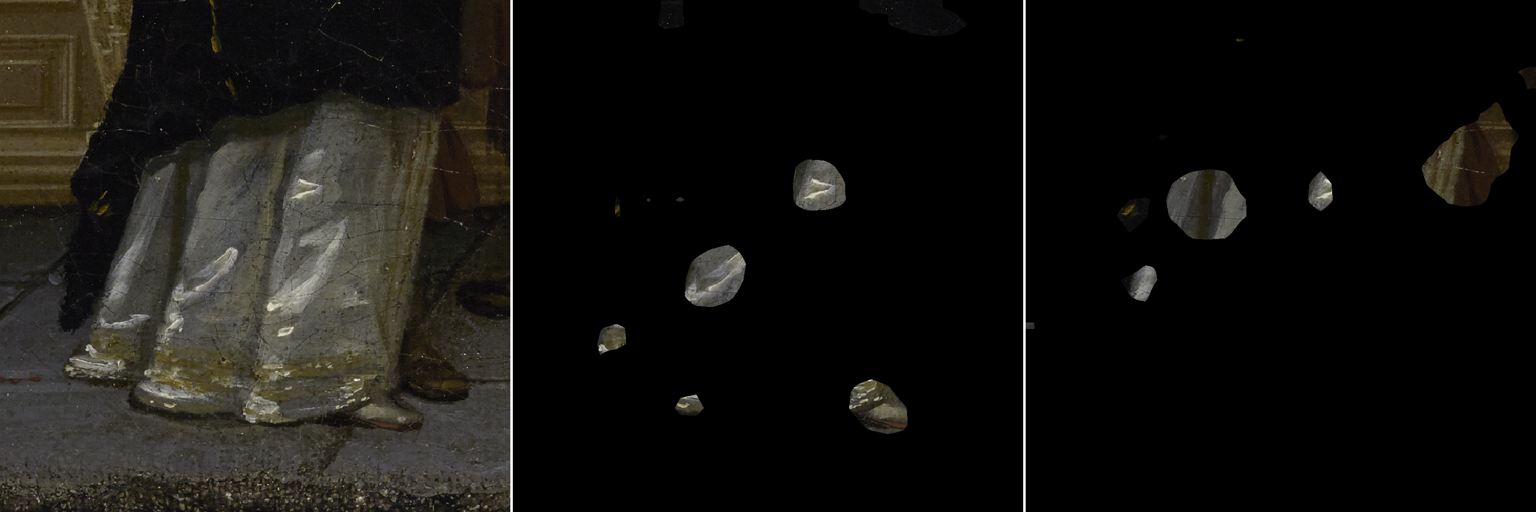}
      \vspace{-4mm}

      \caption{ \textbf{Classifier Cues.} Left to Right: Original Image, Masked Image
      (Painting Classifier), and Masked Image (Photo Classifier). The unmasked
      regions represent evidence used by the classifiers for predicting
      ``silk/satin'' in this particular image. }
      \vspace{-10mm}

    \label{fig:gradcam}
    \end{center}
\end{figure}

\paragraph{Interpretability of Classifier Cues.} Are the cues used by each
classifier interpretable to humans? We produce evidence heatmaps with GradCAM
\cite{gradcam} from the feature maps in the network before the fully connected
classification layer.  We extract high resolution feature maps from images of
size 1024 $\times$ 1024 (for a feature map of size 32 $\times$ 32). The heatmaps
produced by GradCAM show which regions of an image the classifier uses as
evidence for a specific class. If a classifier has learned a good
representation, the evidence that it uses should be more interpretable for
humans. For both models, we compute heatmaps for test images corresponding to
their ground truth label. We conduct a user study on Amazon Mechanical Turk to
find which heatmaps are preferred by humans. Users are shown images with regions
corresponding to heatmap values that are above 1.5 standard deviations above the
mean. Fig \ref{fig:gradcam} illustrates an example. Our user study resulted in
responses from 85 participants, 57 of which were analyzed after quality control.
For quality control, we only kept results from participants who spent over 1
second on average per task item.

\subsubsection{Results. } We find that the classifier trained with paintings
exhibits better cross-domain generalization, and uses cues that humans prefer
over the photo classifier. This suggests that paintings can improve the
robustness of classifiers for this task of fabric classification.

\paragraph{Generalizability of Classifiers.} In Table \ref{table:fg_fabric}, the
performance of the two classifiers are summarized. We find that both classifiers
perform similarly well on the domain they are trained on. However, when the
classifiers are tested on cross-domain data, we find that the painting-trained classifier
performs better than the photo-trained classifier.  This suggests that the
classifier trained on paintings has learned a more generalizable feature
representation for this task.

\begin{figure}[h!]
    \begin{center}
      \vspace{-5mm}
      \includegraphics[width=0.7\linewidth]{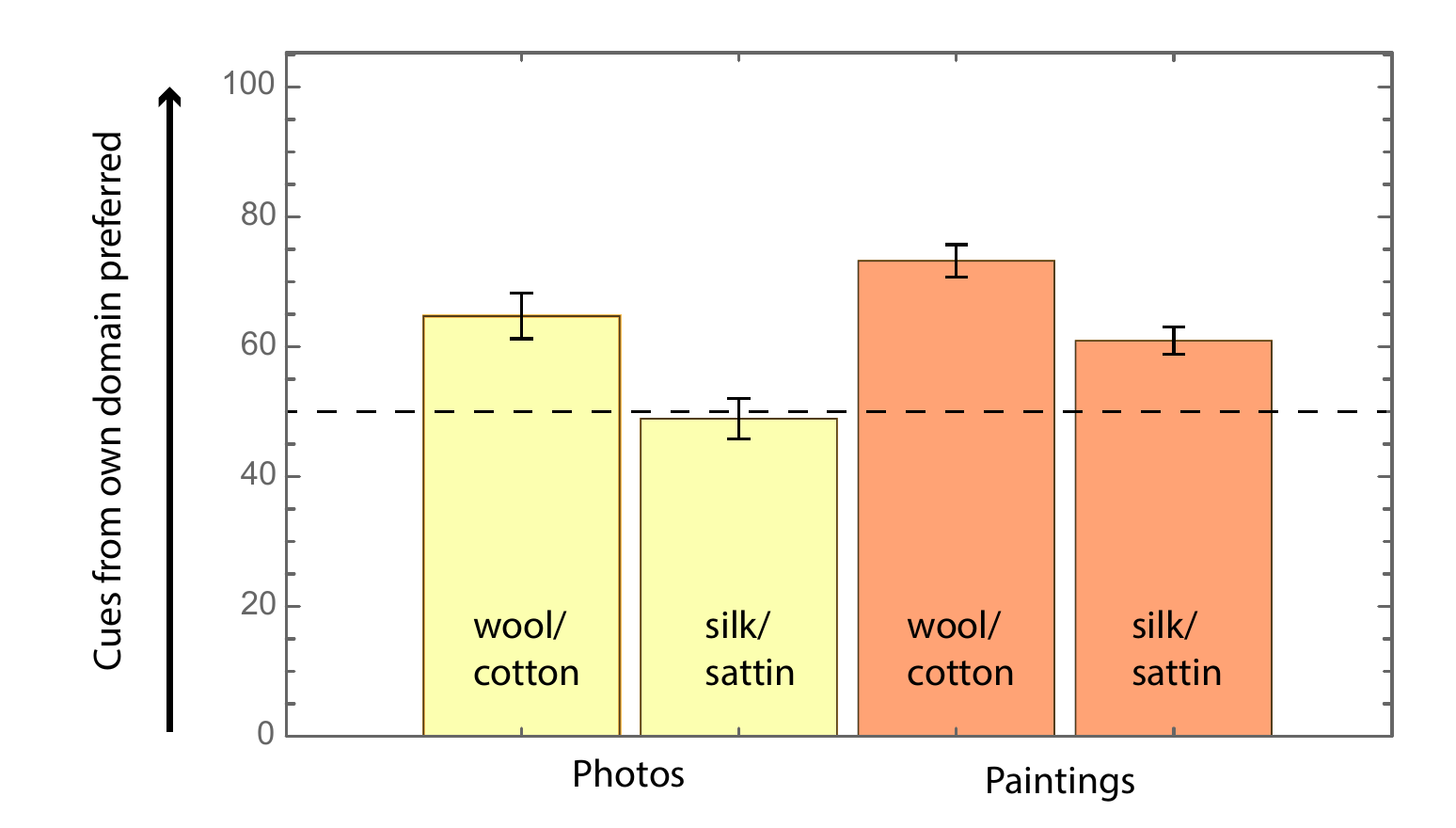}
      \vspace{-4mm}

      \caption{ \textbf{Human Agreement with Classifier Cues.} Pictorial
      representation of user study results from Table \ref{table:fg_user_study}.
      The y-axis represents how often humans prefer the cues from a classifier
      trained on the same domain as the test images. It is clear that humans
      prefer the painting classifier for paintings more than they prefer the
      photo classifier for photos. Interestingly, the painting and photo
      classifiers are equally preferred for silk/satin photos despite
      the painting classifier never seeing a photo during training (bar 2).}

      \label{fig:fg_user_study_figure}
    \end{center}
\end{figure}

\paragraph{Interpretability of Classifier Cues.}  Overall, we find that the
classifier trained on paintings uses evidence that is better aligned with
evidence preferred by humans (Table \ref{table:fg_user_study} and Fig.
\ref{fig:fg_user_study_figure}). Due to domain shifts when applying classifiers
to out-of-domain images, we would expect the cues selected by the painting
classifier to be preferable on paintings, and the cues selected by the photo
classifier to be preferable on photos. Interestingly, this does not hold for
photos of satin/silk (column 2 of Table \ref{table:fg_user_study}) -- we find
that users equally prefer the evidence selected by the painting classifier to
the evidence selected by the photo classifier.  This suggests that either (a)
the painting classifier has learned the ``correct'' human-interpretable cues for
recognizing satin/silk, or (b) that the photo classifier has learned to classify
satin/silk based on some spurious contextual signals.  We asked users to
elucidate their reasoning when choosing which set of cues they preferred. In
general, users noted that they preferred the network which picks out regions
containing the target class. Therefore, it seems that the network trained on
paintings has learned better to distinguish fabric through the actual presence
of such fabrics in the image over other contextual signals.

Taken together, our results provide evidence that a classifier trained on
paintings can be more robust than a classifier trained on photographs. It would
be interesting to explore this further.  A limitation of this study is the
relatively small number of data samples, and very limited number of material
types (two: cotton/wool and silk/satin) that we explored. Are there other
materials or objects which deep neural networks can learn to recognize better
from paintings than photographs?

\begin{table}[h!]
    \centering
    \begin{tabularx}{0.7\linewidth}{|l||X|X|}
    \hline
      & \makecell[l]{Photo $\rightarrow$ Photo} & \makecell[l]{Painting$\rightarrow$ Painting} \\
    \hline
    \makecell[l]{MEAN F1 Score} & 79.6\% & 80.5\% \\
    \hline
    \hline
      & \makecell[l]{Photo $\rightarrow$ Painting} & \makecell[l]{Painting$\rightarrow$ Photo} \\
    \hline
    \makecell[l]{MEAN F1 Score} & 49.5\% & 57.8\% \\

    \hline 
    \end{tabularx}

    \caption{ \textbf{Classifier Generalization}. Classifiers are trained to
    distinguish cotton/wool from silk/satin.  One classifier is trained on
    photographs and another classifier is trained on paintings. Both classifiers
    perform similarly well on images of the same type they were trained on, but
    the classifier trained on paintings performs better on photographs than vice
    versa. This suggests that the features learned from paintings are more
    generalizable for this task on this set of data.}
    \vspace{-10mm}

    \label{table:fg_fabric}
\end{table}

\begin{table}[h!]
    \centering
    \begin{tabularx}{\linewidth}{|l||l|l|l|l|X|}
    \hline
      & \makecell[l]{Cotton/Wool\\Photos} & \makecell[l]{Silk/Satin\\Photos} &
      \makecell[l]{Cotton/Wool\\Paintings} & \makecell[l]{Silk/Satin\\Paintings}
      &\makecell[l]{MEAN} \\
    \hline
      \makecell[l]{Photo Classif.\\Preferred} &
      \makecell[l]{64.7 $\pm$ 3.5\%}
      & \makecell[l]{48.9 $\pm$ 3.1\%}& \makecell[l]{26.8 $\pm$
      2.5\%}&
      \makecell[l]{39.1 $\pm$ 2.1\%}& \makecell[l]{44.9 $\pm$
      1.9\%}\\
    \hline 
      \makecell[l]{Painting Classif.\\Preferred} &
      \makecell[l]{35.3 $\pm$ 3.5\%}
      & \makecell[l]{51.1 $\pm$ 3.1\%}& \makecell[l]{73.2 $\pm$
      2.5\%}&
      \makecell[l]{60.9 $\pm$ 2.1\%}& \makecell[l]{55.1 $\pm$
      1.9\%}\\
    \hline 
    \end{tabularx}

    \caption{ \textbf{Human Agreement with Classifier Cues.} On average, humans
    prefer the cues used by the painting-trained classifier to make its
    predictions over the cues used by the photo-trained classifier.
    Interestingly, the human judgements also indicate that the painting-trained
    classifier uses cues that are just as good to the cues used by the
    photo-trained classifier for silk/satin photos despite never seeing a
    silk/satin photo during training (column 2). A pictorial representation of
    the results is given in Fig. \ref{fig:fg_user_study_figure}.  }

    \label{table:fg_user_study}
\end{table}

\subsection{Benchmarking Unsupervised Domain Adaptation}

In unsupervised domain adaptation (UDA), models are trained on a `source'
dataset with annotated labels as well as an unlabeled `target' dataset. The goal
is to train a model which performs well on unseen target dataset samples.
Existing domain adaptation benchmark datasets for classification focus primarily
on object recognition and tend to be limited in number of data samples, with
most class categories containing on the order of 1000 samples or fewer (for
example, refer to Table 1 of \cite{domainnet}). In contrast, the dataset we use
here has the unique properties of (a) focusing primarily on material
classification and (b) containing on the order of 10-30K for 9 of the 15
annotated classes (e.g. fabric, wood), with the remainder in the range of 2K-5K
(e.g. ground). This positions this data as a valuable addition for benchmarking
for UDA algorithms.

\subsubsection{Experimental Setup.} For this study, we focus on a family of
domain adaptation algorithms which aim to explicitly minimize feature
discrepancy across the source and target domains.  Existing work has shown that
class-conditional UDA in which labels are estimated for target domain samples
during training can be better than class-agnostic UDA where adaptation is
performed without using any estimated label information at all.  We choose CDD
\cite{CDD} and MMD \cite{MMD,tzeng2014deep} as representative methods for
class-conditional and class-agnostic discrepancy minimization.  All methods are
trained with default settings from publicly available source code for CDD, which
includes the use of domain batch normalization \cite{DBN}. We selected 10
material categories: ceramic, fabric, foliage, glass, liquid, metal, paper,
skin, stone, and wood. For our painting dataset, we sampled
as-class-balanced-as-possible from these classes to form a dataset with 10K
samples and a dataset with 60K samples. A corresponding photograph dataset is
constructed from Opensurfaces/MINC/COCO with 10K and 60K samples as well.

\subsubsection{Results.} 

We find that the studied domain adaptation algorithms can indeed behave
differently than would be expected from results on existing benchmark datasets.
This is could due to more data being available or a more difficult domain shift
than conventional adaptation benchmark datasets.

\paragraph{Effect of Dataset Size.} Results are summarized in Table
\ref{table:UDA}.  With the conventional 1K samples per class, we confirm domain
adaptation yields gains over source-only as found on existing benchmark
datasets. In contrast to results on existing benchmarks however, we find that
class-conditional adaptation does not necessarily outperform class-agnostic
adaptation. We hypothesize this occurs due to failures in target label
estimation for the class-conditional case -- we discuss this further below.
Next, with 6K samples per class (which is 6$\times$ more data samples per class
than conventional UDA benchmarks), we find that source-only (i.e., no
adaptation) performs very competitively. In fact, source-only strictly
outperforms adaptation for Painting to Photo transfer in this data regime! This
result suggests that domain adaptation is useful in lower data regimes, but
source-only is a competitive alternative when more data is available. We leave
a deeper exploration of this phenomenon to future work.

\paragraph{Effect of Class Label Estimation.} As found above, class-conditional
adaptation can underperform class-agnostic adaptation despite utilizing more
information. As class-conditional adaptation depends on estimated target labels,
large domain shifts that hamper label estimation can harm adaptation.  To
confirm this, we consider two experiments: CDD with intraclass discrepancy
minimization only (instead of both intraclass minimization and interclass
maximization), and CDD with ground truth labels (i.e., perfect label
estimation). Results are in Table \ref{table:UDA_intra}. In both cases, we see
performance improves.  In the case where perfect label estimation is assumed,
then CDD does outperform intraCDD and MMD as found on existing datasets.
Therefore, estimating class labels for domain adaptation is useful in practice,
but only if the labels are estimated sufficiently well.

\begin{table}[ht!]
    \centering
    \begin{tabularx}{0.8\linewidth}{|l||X|X|}
    \hline
    \makecell[c]{\textbf{ResNet18} \\\emph{1K imgs/class per domain}} & \makecell[l]{Photo $\rightarrow$ Painting} & \makecell[l]{Painting $\rightarrow$ Photo} \\
    \hline
    \makecell[c]{Source-Only} & 35.2 & 46.9 \\
      \makecell[c]{MMD} & 46.1 (+10.9) & 56.4 (+9.5) \\
      \makecell[c]{CDD} & 40.5 (+5.3) & 57.4 (+10.5) \\
    \hline
    \hline
    \makecell[c]{\textbf{ResNet18} \\\emph{6K imgs/class per domain}} & \makecell[l]{Photo $\rightarrow$ Painting} & \makecell[l]{Painting $\rightarrow$ Photo} \\
    \hline
    \makecell[c]{Source-Only} & 38.7 & 53.6 \\
      \makecell[c]{MMD} & 43.5 (+4.8) & 51.5 (-2.1) \\
      \makecell[c]{CDD} & 35.6 (-3.1) & 49.4 (-4.2) \\
    \hline
    \end{tabularx}

    \caption{\textbf{Effect of Dataset Size.} UDA from photo (source) to
    painting (target) and painting (source) to photo (target).  Source-only
    refers to a reference baseline where no adaptation is used.  The gap between
    source-only and UDA decreases as data samples increases from 1K images per
    class to 6K images per class. Furthermore, in contrast to behavior found on
    existing benchmark datasets, the class-conditional method of CDD does not
    necessarily outperform the class-agnostic counterpart MMD. }

    \label{table:UDA}
\end{table}

\begin{table}[h!]
    \vspace{-5mm}
    \centering
    \begin{tabularx}{0.8\linewidth}{|l||X|X|}
    \hline
    \makecell[c]{\textbf{ResNet18} \\\emph{1K imgs/class per domain}} & \makecell[l]{Photo $\rightarrow$ Painting} & \makecell[l]{Painting $\rightarrow$ Photo} \\
    \hline
    \makecell[c]{MMD} & 46.1 (+10.9) & 56.4 (+9.5) \\
    \makecell[c]{IntraCDD} & 44.4 (+9.2) & 58.5 (+11.6) \\
    \makecell[c]{CDD} & 40.5 (+5.3) & 57.4 (+10.5) \\
    \hline
      \makecell[c]{IntraCDD w/ GT labels} & 57.6 (+22.4) & 64.4
      (+17.5) \\
      \makecell[c]{CDD w/ GT labels} & 61.6 (+26.4) & 72.5 (+25.6) \\
    
    \hline 
    \end{tabularx}

    \caption{ \textbf{Effect of Class Label Estimation.} Reducing the reliance
    class label estimation improves class-conditional UDA when label
    estimation for target data is poor. We find that IntraCDD (which considers
    only intraclass discrepancy) outperforms CDD (which considers both
    intraclass and interclass discrepancy).  Using ground truth (GT) labels with
    CDD (i.e., assuming perfect class label estimation) recovers performance
    gains over intraCDD and MMD. MMD does not require class label estimation,
    and so its performance not suffer in the case of poor label estimation.}

    \label{table:UDA_intra}
\end{table}

\section{Conclusion}

In this paper, we explored how modern deep learning tools developed for natural
images can be used to analyze paintings, and in turn, how paintings can be used
to improve deep learning systems in a series of experiments. Our findings
suggest that progress in visual perception for natural images can benefit
systems used for fine art analysis, and having access to the visual information
encoded in paintings can be fruitful for building more generalizable perception
systems.

\section{Acknowledgements}

This work was funded in part by NSF (CHS-1617861 and CHS-1513967), NSERC
(PGS-D 516803 2018), and the Netherlands Organization for Scientific Research
(NWO) project 276-54-001.

%
%
%
\bibliographystyle{splncs04}
\bibliography{paper}
\end{document}